\documentclass[letterpaper, 10 pt, conference]{ieeeconf}  

\IEEEoverridecommandlockouts                           

\usepackage{amsmath, amssymb , graphicx, subcaption}
\usepackage{multirow}
\usepackage{dblfloatfix} 
\usepackage{balance}
\usepackage{hyperref}
\DeclareCaptionFont{caption}{\fontsize{8}{9.6}\selectfont}
\title{\LARGE \bf
CRANE: A highly dexterous needle placement robot for evaluation of interventional radiology procedures}

\author{Dimitri A. Schreiber$^1$, Hanpeng Jiang$^{\dag,1}$, Guosong Li$^{\dag,1}$, Julie Yu$^{\dag,2}$, Zhaowei Yu$^{\dag,1}$, Renjie Zhu$^{\dag,1}$,\\ Alexander M. Norbash$^3$, and Michael C. Yip$^{1}$, \IEEEmembership{Member, IEEE}
\thanks{$^\dag$ Equal contributions}
\thanks{$^1$Department of Electrical and Computer Engineering, University of California San Diego, La Jolla, CA 92093 USA. {\tt\small \{dschreib, hjiang, g4li, zhy125, rezhu, m1yip\}@eng.ucsd.edu}}
\thanks{$^2$Department of Mechanical and Aerospace Engineering, University of California San Diego, La Jolla, CA 92093 USA. {\tt\small jhy015@ucsd.edu}}
\thanks{$^3$Department of Radiology, University of California San Diego, La Jolla, CA 92093 USA. {\tt\small anorbash@ucsd.edu}}}

\begin{document}

\maketitle
\thispagestyle{empty}
\pagestyle{empty}

\begin{abstract}
Interventional Radiology (IR) enables earlier diagnosis and less invasive treatment of numerous ailments. Here we present our ongoing development of CRANE: CT Robotic Arm and Needle Emplacer, a robotic needle positioning system for CT guided procedures. 
The robot has 8 active Degrees-of-Freedom (DoF) and a novel infinite travel needle insertion mechanism.
The control system is distributed using the Robot Operating System (ROS) across a low latency network that interconnects a real-time low-jitter controller, with a desktop computer which hosts the User Interface (UI) and high level control.
This platform can serve to evaluate limitations in the current procedures and to prototype potential solutions to these challenges in-situ.
\end{abstract}

\section{Introduction}
CT guided needle interventions are widely used and play an essential role within the field of Interventional Radiology (IR). 
Percutaneous core-needle biopsy of pulmonary nodules and injections for lumbosacral spine pain are two of the most common IR procedures. 
Primary lung cancer is the leading cause of cancer death and accounts for approximately $13\%$ of all new cancer cases, worldwide \cite{cancerorg}. 
Definitive diagnosis requires direct tissue samples, frequently acquired via core-needle biopsy. 
However, physicians have challenges reaching masses under $1cm$ in diameter \cite{Tian2017} and even struggle with larger masses when far from the surface \cite{Yankelevitz1996}. 
Separately, lumbo-sachrial spine is the area of the body most frequently responsible for chronic pain\cite{Gangi1998}.
Needle injections of steroids, nerve-blocking agents, and coagulants show promise for short and long term pain relief. 
Due to the complex 3D geometry of the spine and the precise needle placement required for effective pain relief, hand-guided needle placement is difficult. 
These goals of diagnosis and pain reduction are readily complicated by current procedural challenges, including the repetitive translation of the patient into and of the bore for CT scans in-between reference-free manual adjustment of the needle.

A general-purpose robotic system will enable decreased patient radiation exposure through fewer CT scans, fewer needle insertions with increased freedom in the needle approach angle.

\begin{figure}[t!]
    \centering
      \includegraphics[width=\linewidth,trim={4cm 0 4cm 0},clip=true]{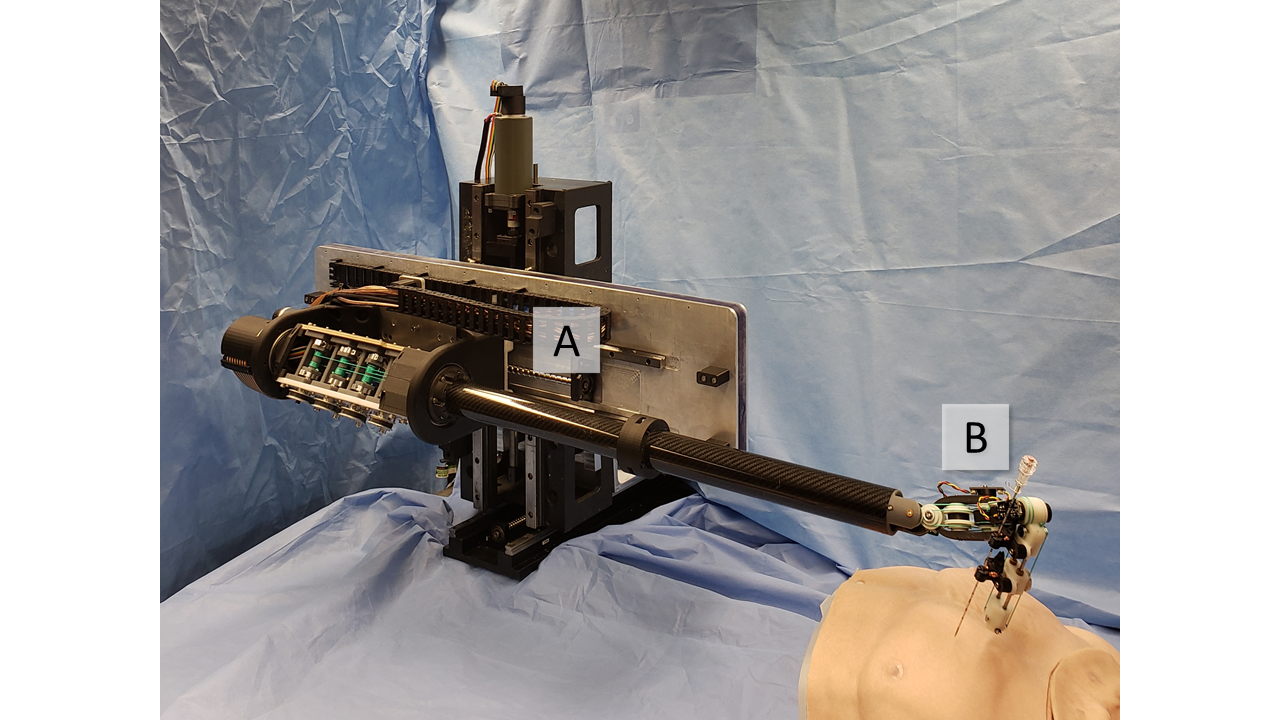}
    \caption{CT-guided robotic system with a low-intra-bore-profile serial-link redundant 8-DoF arm.  \boxed{A} is the back-end positioning stage.  \boxed{B} is the remote 4-DoF arm and clutching needle driver.}
    \label{fig:robot_person_UCSD}
\end{figure}

\begin{figure*}[t!]
    \centering
      \includegraphics[width=0.92\linewidth]{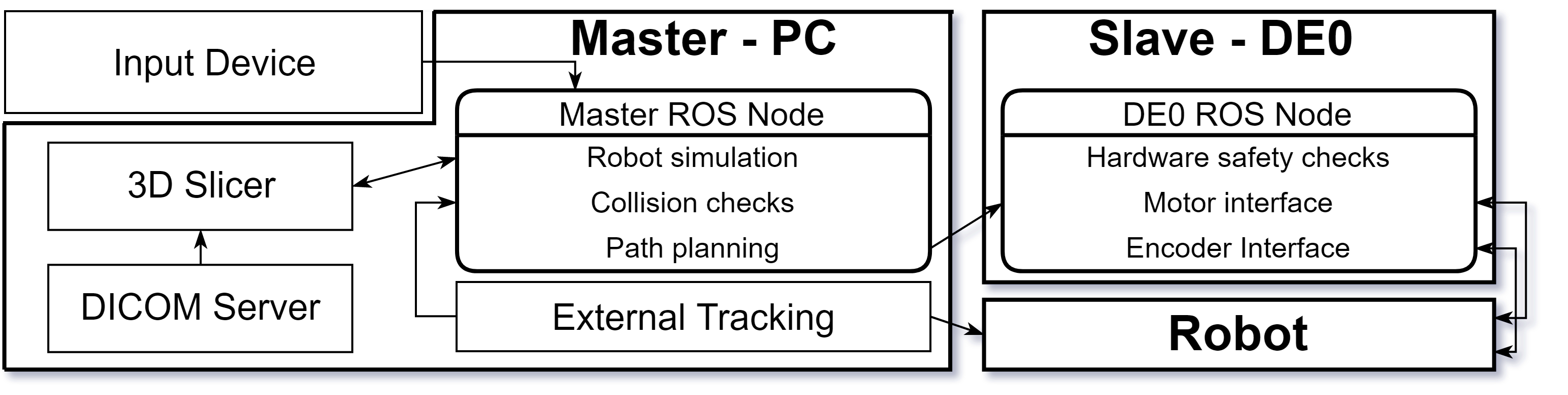}
    \caption{The robot simulation keeps track of the simulated robot configuration and reports that to the external medical software for a robot overlay on the DICOM images. This combination helps the physician more accurately choose and plan a setup needle pose. The collision checks and the path planning work towards an initial robot configuration and giving instructions for the actual robot movement.}
    \label{fig:system_diagram}
\end{figure*}

\section{System Overview}
\subsection{Mechanical Design}
Both the mechanics of the system have been greatly improved, and the kinematics have been modified from our previous work \cite{Schreiber2019} to better suit IR needle placement procedures. Here we present an 8 DoF system composed of two distinct structures: 4 DoF intra-bore with needle driving clutches and 4 DoF exo-bore.

The intra-bore structure's serial-link arm is cable-driven by motors located on the exo-bore structure.
Improvements have been made to the remote drive and arm,
including: 
\begin{enumerate}
    \item lower gear ratio motors for faster motion and inherent safety via torque-sensing,
    \item decreasing link lengths to 7cm and increasing joints' range of motion for improved dexterity,
    \item incorporation of joint mounted high resolution absolute magnetic encoders for improved system state tracking,
    \item development of a novel clutch mechanism using Shape Memory Alloy actuators helicaly wrapped around a flexure, allowing infinite depth needle insertion and retraction with a short insertion stage.
\end{enumerate}
The exo-bore backend, composed of a revolute axis on the trunnion and an X-Y-Z cartesian stage driven by ballscrew actuators, allows for incredibly precise and accurate motion. 
\subsection{Electrical Hardware}
The low-level controller is based on our previous design \cite{Schreiber2019}. However, the system has been significantly improved in its reliability, ease of use, and extensibility. This system is composed of a System-on-Chip Arm Cortex A9 combined with a Cyclone V Field Programmable Gate Array running Debian Linux performing synchronous Proportional Integral Derivative motor position control at 1khz on all axis. 
Hardware and software safety interlocks are incorporated, including several watchdog timers, heartbeats, and a physical emergency stop switch. 
This low-level controller hosts a ROS node to allow the remote master PC to update the motor position setpoints and disable/enable the robot, as well as updating the remote master PC with position feedback from the magnetic encoder. 
The needle clutch temperature controller is implemented using a PSoC MCU with analog clutch driver circuits, with thermistors for temperature feedback.
\subsection{Software Systems}
ROS is used for inter-system communication. 
On the master PC (see Fig. \ref{fig:system_diagram}), there are three main components: a master ROS node, a DICOM viewer (3D Slicer),  and an external robot tracking software. The master ROS node consists of several parts, as shown in Fig. \ref{fig:system_diagram}. 
The remote low-level controller's ROS node receives instructions over Ethernet from the remote master node.
The needle clutch controller is implemented through a FreeRTOS port on a PSoC MCU. By utilizing the Real-Time Operating System's deterministic nature, this setup allows exact control and robust operation of the clutches.
\subsection{Clinical Workflow}
\begin{figure}[b!]
    \centering
      \includegraphics[width=0.92\linewidth]{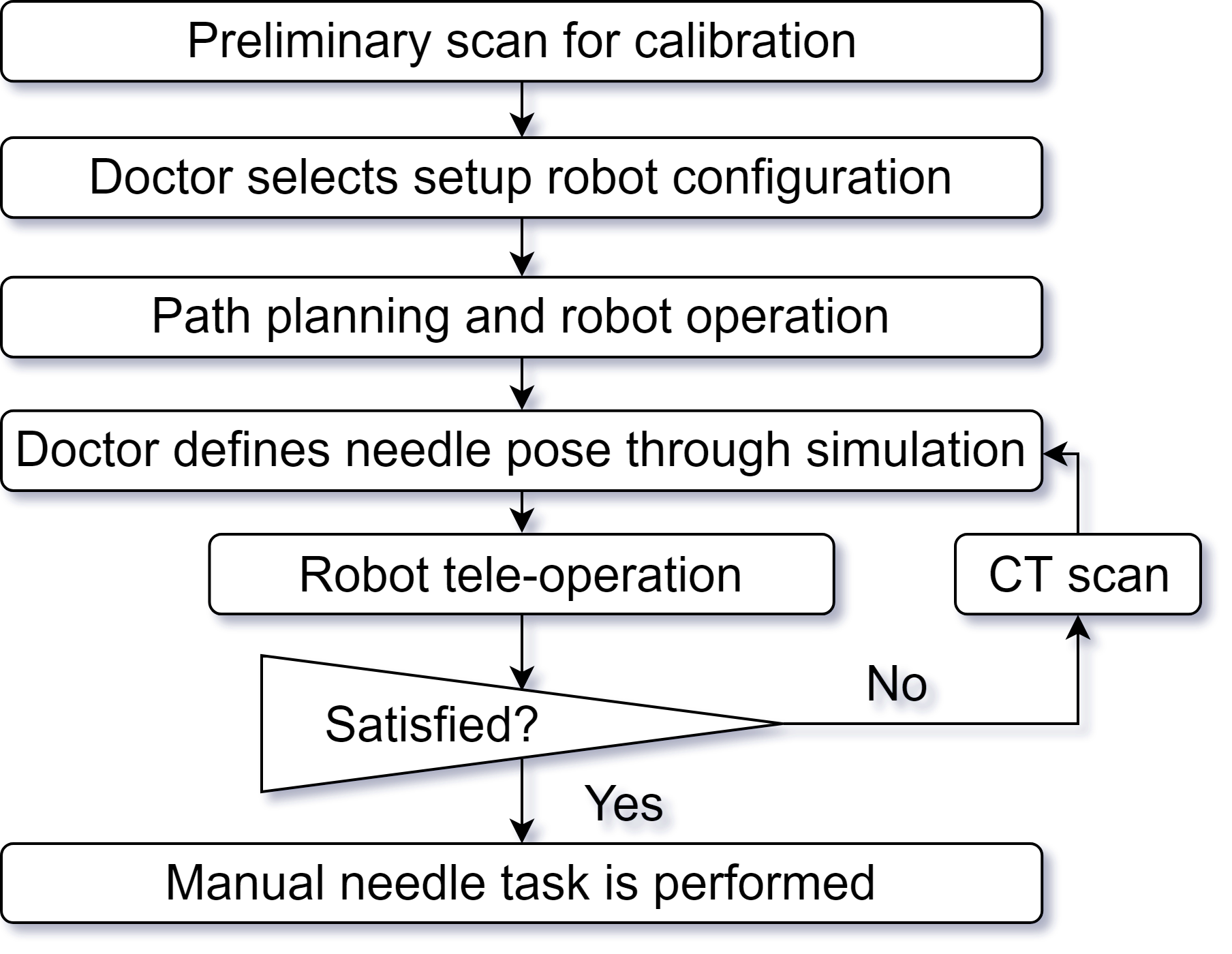}
    \caption{Clinical workflow for the robotic needle positioning system.}
    \label{fig:workflow}
\end{figure}

The system's clinical workflow is illustrated in Fig. \ref{fig:workflow}. 
First, the robot to scanner calibration transform is calculated using the preliminary scan. Then, the physician manipulates the needle into the desired setup pose. Once this setup pose is confirmed, the robot configuration is optimized to maximize the manipulability, distance from a collision, and joint limits while maintaining the correct needle pose and offering a collision-free path from the initial configuration. Following physician confirmation, the robot follows this trajectory to the setup pose. The physician now iterates between simulated needle motions, robot teleoperation, and CT confirmation scans until the target is reached. Upon satisfactory positioning, the physician proceeds with the manual needle biopsy or injection procedure.
\section{Conclusion and Future work}
Here we present a robust robotics platform for IR needle procedures with a novel clutching needle driver. This system is highly dexterous and has a fully integrated software system with a DICOM viewer and a 3D UI, redundant sensors, and safety systems. Future work will explore clinical procedures, where the system is used to evaluate limitations in current procedures and solutions are proposed, developed, and evaluated. Also, a more critical analysis of the system will be presented in later work.

\balance
\bibliographystyle{ieeetr}
\bibliography{iros2019_workshop_ct_robot}

\end{document}